\documentclass{article}

\usepackage{arxiv}

\usepackage[utf8]{inputenc} 
\usepackage[T1]{fontenc}    
\usepackage{hyperref}       
\usepackage{url}            
\usepackage{booktabs}       
\usepackage{amsfonts}       
\usepackage{nicefrac}       
\usepackage{microtype}      
\usepackage{lipsum}
\usepackage{graphicx}
\usepackage{graphicx}
\usepackage{comment}
\usepackage{amsmath,amssymb}
\usepackage{color}
\usepackage{multirow,array, bigdelim, makecell, booktabs}
\usepackage{slashbox}
\usepackage{multirow}
\usepackage{booktabs}
\usepackage{rotating}
\usepackage{xcolor,colortbl}
\usepackage{hyperref}
\graphicspath{ {./images/} }

\DeclareRobustCommand{\eg}{\emph{e.g. }}

\DeclareRobustCommand{\etal}{{et al. }}

\DeclareRobustCommand{\ds}{{QMAR}}
\DeclareRobustCommand{\ki}{{KIMORE}}

\title{{VI-Net: {View-Invariant} Quality of Human Movement Assessment}}

\author{
 Faegheh Sardari \\
  Department of Computer Science\\
  University of Bristol\\
  Bristol, UK \\
  \texttt{faegheh.sardari@bristol.ac.uk} \\
   \And
 Adeline Paiement \\
  Laboratoire d'Informatique et Syst\`{e}mes\\
  University of Toulon\\
  Toulon, France \\
  \texttt{adeline.paiement@univ-tln.fr} \\
  \And
 Sion Hannuna \\
  Department of Computer Science\\
  University of Bristol\\
  Bristol, UK \\
  \texttt{sh1670@bristol.ac.uk} \\
  \And
  Majid Mirmehdi \\
  Department of Computer Science\\
  University of Bristol\\
  Bristol, UK \\
  \texttt{m.mirmehdi@bristol.ac.uk} \\
}

\begin{document}
\maketitle
\begin{abstract}
We propose a view-invariant method towards the assessment of the quality of human movements {which does 
 not rely on}  skeleton data.  Our end-to-end convolutional neural network consists of two stages, where at first a view-invariant trajectory descriptor for each body joint is generated from RGB images, and then the collection of trajectories for all joints are processed by an adapted, pre-trained 2D CNN (e.g. VGG-19 or ResNeXt-50) to learn the relationship amongst the different body parts and deliver a score for the movement quality. We release the only publicly-available, multi-view, non-skeleton, non-mocap, rehabilitation movement dataset (\ds), and provide results for both cross-subject and cross-view scenarios on this dataset. We show that  VI-Net achieves average rank correlation of {0.66} on cross-subject and {0.65} on unseen views when trained on only two views. We also evaluate the proposed method on the single-view rehabilitation dataset KIMORE and obtain 0.66 rank correlation against a baseline of 0.62. 
\end{abstract}

\section{Introduction} \label{sec:intro}
{Beyond the realms of action detection and recognition, action analysis includes the automatic assessment of the quality of human action or movement, e.g. in} sports action analysis 
\cite{li2018end,parmar2017learning,parmar2019and,pan2019action}, skill assessment \cite{fard2018automated,Doughty_2019_CVPR}, and patient rehabilitation movement analysis \cite{sardari2019view, liao2019deep}. 
For example, in the latter application,
clinicians observe patients performing specific actions in the clinic, such as walking or sitting-to-standing, to establish {an objective marker for their  level of functional mobility.}  
By automating such mobility disorder assessment using computer vision, 
health service authorities can decrease costs, reduce hospital visits, and diminish the variability in clinicians' subjective assessment of patients.

{Recent  RGB based {action analysis} methods, such as \cite{parmar2017learning,parmar2019and,Doughty_2019_CVPR,pan2019action}, are not able to deal with view-invariance when applied to viewpoints significantly different to their training data}. 
{To achieve some degree of invariance, some works} 
such as \cite{crabbe2015skeleton,liao2019deep,sardari2019view,khokhlova2019normal,antunes2018aha,blanchard2019keep,lei2019survey}, have made use of 3D human pose obtained from (i) Kinect, (ii) motion capture, or (iii) 3D pose estimation methods. Although the Kinect can provide 3D pose efficiently in optimal conditions, it is dependent on several parameters, 
including distance and viewing direction between the subject and the  sensor. Motion capture systems (mocaps) tend to be highly accurate and view-invariant, but {obtaining} 3D pose by such means is expensive and time consuming, since it requires specialist hardware, software, and setups. 
These make mocaps unsuitable for use in unconstrained home or clinical or sports settings.
Recently, many deep learning methods, e.g. \cite{wandt2019repnet,zhao2019semantic,zhou2019hemlets,kolotouros2019learning, kocabas2020vibe}, have been proposed to extract 3D human pose from RGB images. {Such methods (a) either do not deal with  view-invariance and are trained from specific views on their respective datasets (for example,  \cite{wandt2019repnet,kolotouros2019learning} show that their methods fail when they apply them on poses and view angles which are different from their training sets), (b) or if they handle view-invariance, such as \cite{qiu2019cross,remelli2020lightweight}, then they need multiple views for training.}

\begin{figure}[t]
\centering
    \includegraphics[width=1\linewidth]{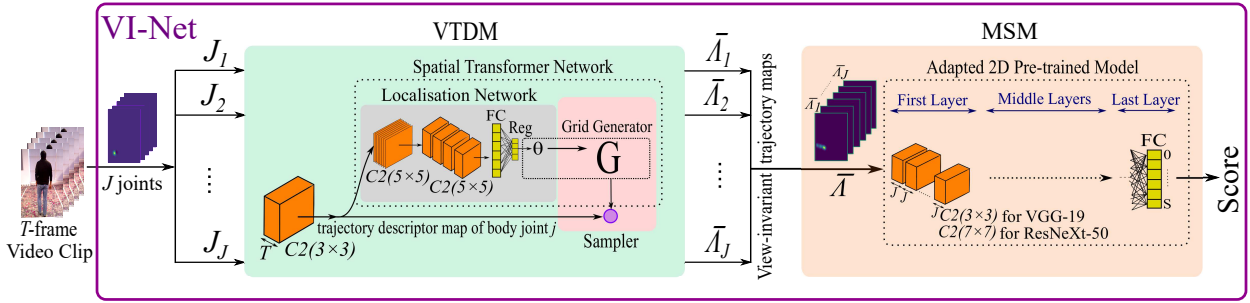}
    \vspace*{-5mm}
       \caption{{VI-Net has an view-invariant trajectory descriptor module (VTDM) and a movement score module (MSM) where the classifier output corresponds to a quality score.}}
    \label{fig:overal schema}
\end{figure}

{To the best of our knowledge, there is no existing {\it RGB-based, view-invariant} method that assesses the quality of human movement.} We argue here that using temporal pose information from RGB, can be repurposed, instead of skeleton points, for {view-invariant} movement quality assessment. In the proposed end-to-end View-Invariant Network (VI-Net in Fig.~\ref{fig:overal schema}), we  stack temporal heatmaps of each body joint (obtained from OpenPose \cite{cao2017realtime}) and feed them into our view-invariant trajectory descriptor module  (VTDM). This applies a 2D convolution layer that aggregates spatial poses over time to generate a trajectory descriptor map per body joint, which is then forged to be view-invariant by deploying the Spatial Transformer Network \cite{jaderberg2015spatial}.  Next, in our movement score module (MSM), these  descriptor {maps} for all body joints are put through {an adapted} pre-trained 2D convolution model, such as 
VGG-19 \cite{simonyan2014very} or ResNeXt-50 \cite{xie2017aggregated},
to {learn} the relationship amongst the joint trajectories and estimate a score for the movement. {Note, OpenPose has been trained on 2D pose datasets which means that our proposed method implicitly benefits from joint {labelling}.} 

Initially, we apply our method to a new dataset, called  QMAR\footnote{Under preparation for public release.}, that includes multiple camera views of subjects performing both normal movements and simulated Parkinsons and Stroke ailments for walking and sit-to-stand actions. {We provide cross-subject and cross-view results on this new dataset. {Recent works such as \cite{wang2018dividing,li2018unsupervised,lakhal2019view,li2017domain}, provide cross-view results only when their network is trained on multiple views. As recently noted by Varol et al. \cite{varol19_surreact}, a highly challenging scenario} in view-invariant action recognition would be to obtain cross-view results by training from only one viewpoint.} 
While we present results using a prudent set of two viewpoints only within a {multiview training}  scenario, we also rise to the challenge to provide cross-view results by training solely from a single viewpoint. We also present results on the single view rehabilitation dataset KIMORE \cite{capecci2019kimore} {which provides 5 different types of lower back exercises in real patients suffering from Parkinsons, Stroke, and back pain}.

{This work makes a number of contributions.} We propose the first view-invariant method to assess quality of movement from RGB images and our approach does not require any knowledge about viewpoints or cameras during training or testing. Further, it is based on 2D convolutions only 
{which is} computationally cheaper than 3D temporal methods. We also present an RGB, multi-view, rehabilitation movement assessment dataset (\ds) to both evaluate the performance of the proposed method and provide a benchmark dataset for future view-invariant methods.

The rest of the paper is organized as follows. We review the related works in Section \ref{sec:related works} and our \ds~ dataset in Section \ref{sec:dataset}. Then, we present our proposed network in Section \ref{sec:proposed method} and experimental results in Section \ref{sec:experimental results}. Finally, in  Section \ref{sec:concolusion}, we conclude our work. 


\section{Related Work}
\label{sec:related works}
{Action analysis} has picked up relative pace only in recent years with the majority of works covering one of either physical rehabilitation, sport scoring, or skill assessment \cite{lei2019survey}. 
Here, we first consider example non-skeleton based methods (which are mainly on sport scoring), and then review physical rehabilitation methods as it is the main application focus of our work. Finally, given the lack of existing view-invariant movement analysis techniques, we briefly reflect on related  view-invariant action recognition approaches.

{\bf {Non-Skeleton} Movement Analysis - }
{A number of works have focused on scoring sports actions.  Pirsiavash \etal \cite{pirsiavash2014assessing} proposed an SVM based method, trained on spatio-temporal features of body poses, to assess the quality of diving and figure-skating actions. Although their method estimated action scores better than human non-experts, it was less accurate than human expert judgments. }
More recently, deep learning methods have been deployed to assess the quality of sport actions {in RGB-only data}, such as  \cite{li2018end,parmar2017learning,xiang2018s3d,parmar2019and,pan2019action,tang2020uncertainty}. For example, Li et al. \cite{li2018end} 
divided a video into several clips to extract their spatio-temporal features by differently weighted C3D networks and then  concatenated the features 
for input to another C3D network to predict action scores. 
Parmar and Morris presented a new dataset and also
used a C3D network to extract  features for multi-task learning \cite{parmar2019and}. 

The authors in \cite{pan2019action,  tang2020uncertainty} propose I3D \cite{carreira2017quo} based methods to analyse human movement. {Pan \etal \cite{pan2019action} combine I3D features with pose information by building join relation graphs to predict score movement. Tang \etal \cite{tang2020uncertainty} proposed a novel loss function which addresses the intrinsic score distribution uncertainty of sport actions in the decisions by different judges.}  The use of {3D convolutions} imposes a hefty memory and computational burden, even for a relatively shallow model, {which we avoid in our proposed method}. Furthermore,  the performance of these methods {are expected to drop} significantly when they are applied on a different viewpoint since they are trained on appearance features which change drastically in varying viewpoints. 

{\bf {Rehabilitation Movement Assessment - }} Several works have focussed on such movement assessment, e.g.
\cite{crabbe2015skeleton,tao2016comparative,elkholy2019efficient,khokhlova2019normal,sardari2019view, liao2019deep}. {For example,  Crabbe \etal \cite{crabbe2015skeleton} proposed a CNN network to map a depth image to high-level pose in a manifold space made from skeleton data. Then, the high level poses were employed by a statistical model to assess quality of movement for walking on stairs. In \cite{sardari2019view}, Sardari \etal  extended the work in \cite{crabbe2015skeleton} by proposing a ResNet-based model to estimate view-invariant high-level pose from RGB images where the high-level pose representation was derived from 3D mocap data using manifold learning. The accuracy of their proposed method was {{good}} when training was performed from all views, but dropped significantly on unseen views.}

{Liao \etal~\cite{liao2019deep} proposed an LSTM based method for rehabilitation movement assessment from 3D mocap skeleton data and proposed a performance metric based on Gaussian mixture models to estimate their score.} 
{Elkholy \etal \cite{elkholy2019efficient} extracted spatio-temporal descriptors from 3D Kinect skeleton data to assess the quality of movement for walking on stairs, sit-down, stand-up, and walking actions. They first classified each sequence into normal and abnormal by making a probabilistic model from descriptors derived from normal subjects, and then scored an action by modeling a linear regression on spatio-temporal descriptors of movements with different scores. 
Khokhlova \etal \cite{khokhlova2019normal} proposed an LSTM-based method to classify pathological gaits from Kinect skeleton data. They trained several bi-directional LSTMs on different training/validation sets of data. For classification, they computed the weighted mean of the LSTM outputs.}
All the methods that rely on {skeleton} data 
are either unworkable or difficult to apply to in-the-wild scenarios for rehabilitation (or sports or skills) movement analysis.

{\bf View-Invariant Action Recognition - }
As stated in {\cite{rahmani2018learning,li2018unsupervised,varol19_surreact}} amongst others, the performance of action recognition methods, such as  \cite{tran2015learning,carreira2017quo, feichtenhofer2019slowfast,hara2018can,lin2019tsm} to name a few, 
drops drastically  when they test their models from unseen views, {since appearance features change significantly in different viewpoints}. To overcome this, some works have dealt with viewpoint variations through skeleton data, e.g. \cite{ke2017new, liu2017enhanced, rahmani2018learning, zhang2019view}. {For example, Rahmani \etal \cite{rahmani2018learning} train an SVM on view-invariant feature vectors from dense trajectories of multiple views in mocap data via a fully connected neural network.} 
{Zhang \etal \cite{zhang2019view} developed a two-stream method, one LSTM and one convolutional model, where both streams include a view adaptation and  a classification network. In each case, the former network was trained to estimate the transformation parameters of 3D skeleton data to a canonical view, and the latter classified the action.} Finally, {the output of the two streams were fused by weighted averaging of the two classifiers' outputs.}

As providing skeleton data is difficult for in-the-wild scenarios, others such as  \cite{wang2018dividing,li2018unsupervised,lakhal2019view,varol19_surreact,liu2018recognizing} have focused on generating  view-invariant features from RGB-D data. {Li \etal \cite{li2018unsupervised} extract unsupervised view-invariant features by designing a recurrent encoder network which estimated 3D flows from RGB-D streams of two different views. In \cite{varol19_surreact}, the authors generated synthetic multi-view video sequence from one view, and then trained a 3D ResNet-50 \cite{hara2018can} on both synthetic and real data to classify actions.} Among these methods, Varol et al. \cite{varol19_surreact} is the only work that provides cross-view evaluation through single view training, {resulting in $49.4\%$ accuracy on the UESTC dataset \cite{ji2018large}, which then was increased to $67.8\%$ when they used additional synthetic multi-view data for training.}


\begin{figure}[h]
\begin{center}
    \includegraphics[width=0.4\linewidth]{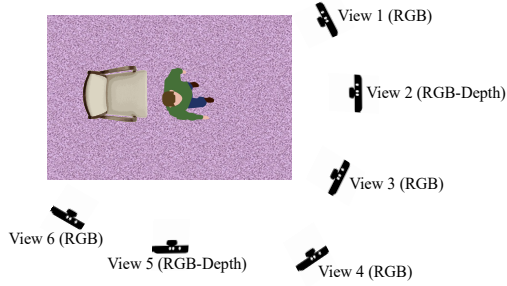}
    \end{center}
       \caption{{Typical camera views in  the {\ds} dataset with each one placed at a different height.}}
    \label{fig:cameras}
\end{figure}
\section{Datasets}    
\label{sec:dataset}
{There are many datasets for healthcare applications, such as \cite{paiement2014online,vakanski2018data,liao2019deep,elkholy2019efficient,capecci2019kimore}, which are single view and only include depth and/or skeleton data}. To the best of our knowledge, there is no existing dataset (bar one) that is suitable for view-invariant  movement assessment from RGB images. The only known multi-view dataset is SMAD, used in Sardari \etal \cite{sardari2019view}. 
{Although it provides RGB data recorded from 4 different views, it only includes annotated data for a walking action and the subjects' movements are only broadly classified into normal/abnormal, without any scores. Thus it is not a dataset we could use for comparative performance analysis.} 

{Next, we first introduce our new RGB multi-view Quality of Movement Assessment for Rehabilitation dataset, \ds. Then, we give the details of a recently released rehabilitation dataset {\ki} \cite{capecci2019kimore}, a single-view dataset that includes RGB images and score annotations, making it suitable for single-view evaluation.}


\subsection{{\ds}}
{\ds} was recorded using 6 Primesense cameras with 38 healthy subjects, 8 female and 30 male. Fig. \ref{fig:cameras} shows the position of the 6 cameras - 3 different frontal views and 3 different side views. The subjects were trained by a physiotherapist to perform two different types of movements while simulating two ailments, resulting in four overall possibilities: {a return walk} to approximately the original position while simulating Parkinsons (W-P), and Stroke (W-S), and standing up and sitting down with Parkinson (SS-P) and  Stroke (SS-S).  The dataset includes RGB and depth (and skeleton) data, {although in this work we only use RGB}. As capturing depth data from the 6 Primesense cameras was not possible due to 
infrared interference, the depth {and skeleton} data {were} retained from only {view 2 at $\approx0^\circ$ and view 5 at $\approx90^\circ$}. 

\begin{figure*}[t]
\begin{center}
    \includegraphics[width=0.98\linewidth]{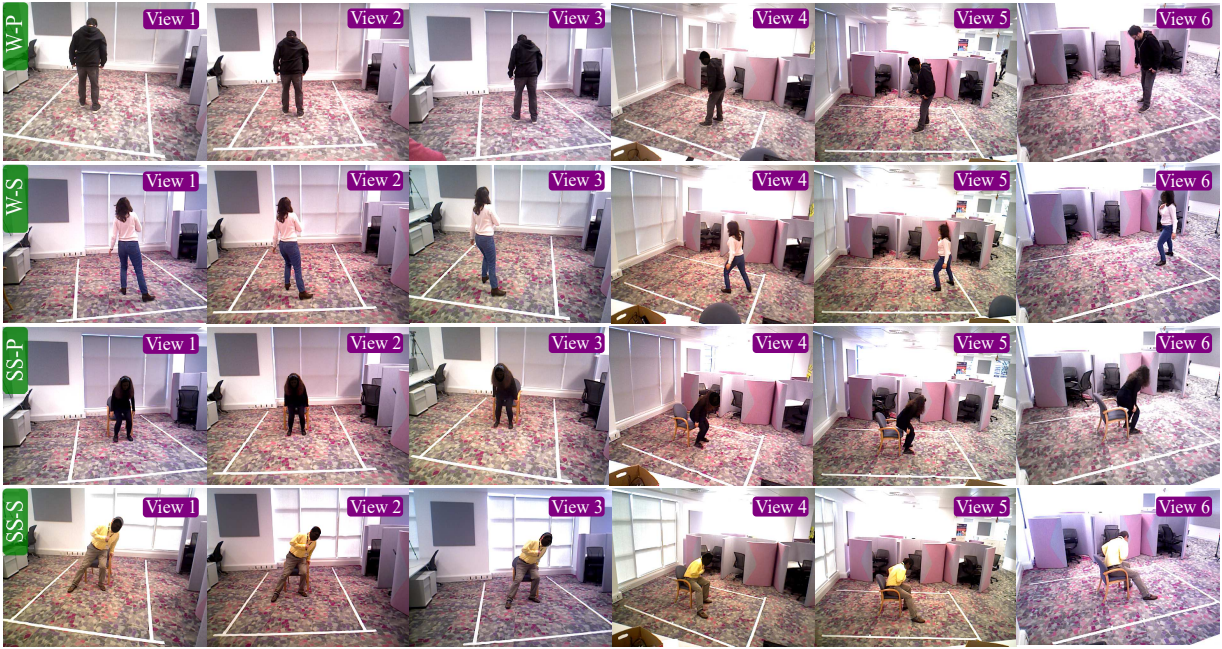}
    \end{center}
    \vspace*{-2mm}
       \caption{Sample frames from {\ds} dataset, showing all 6 views for {(top row) walking with Parkinsons (W-P), (second row) walking with Stroke  (W-S), (third row) sit-stand with Parkinsons  (SS-P), and (bottom row) sit-stand with Stroke.}}
    \label{fig:datset sample}
\end{figure*}

The movements in {\ds} were scored 
by the severity of the abnormality. The score ranges were $0$ to $4$ for W-P, $0$ to $5$ for W-S and SS-S, and $0$ to $12$ for SS-P. A score of $0$ in all cases indicates a normally executed action.
Sample frames from {\ds} are shown in Fig. \ref{fig:datset sample}. {Table \ref{tab:dataset detalis} details the quality score or range and the number of frames and sequences for each action type. Table  \ref{tab:dataset-statistic} details the number of sequences for each score.}

\begin{table}[h]
\centering
\scalebox{0.90}
{
    \begin{tabular}{|c|c|c|c|c|c|c|}  \hline        
    \multicolumn{2}{|c|}{\multirow{2}{*}{\bf Action }} & {\textbf{Quality}} & \multirow{2}{*}{\textbf{\# Sequences}} &{\textbf{\#Frames/Video}}&{\textbf{Total frames}}\\ 
    	\multicolumn{2}{|c|}{} & {\textbf {Score}} &  & Min-Max & \\ \hline \hline
    	{{\bf W}}& {Normal}  & 0   & 41  & 62-179 & 12672\\  \hline 
    	{\bf W-P}& {Abnormal} & 1-4 & 40  & 93-441 & 33618 \\ \hline
    	{\bf W-S}& {Abnormal} & 1-5 & 68  & 104-500 & 57498\\ \hline \hline
    	{{\bf SS}}& {Normal}   & 0   & 42  & 28-132 & 9250\\ \hline 
    	{\bf SS-P}& {Abnormal} & 1-12 & 41  & 96-558 &41808\\ \hline
    	{\bf SS-S}& {Abnormal} & 1-5 & 74  & 51-580 & 47954 \\ \hline
    	
    \end{tabular}}
\caption{{Details of the movements in the {\ds} dataset}.}
\label{tab:dataset detalis}
\end{table}

\begin{table}[h]
    \centering
        \scalebox{0.90}{
            \begin{tabular}{|c|*{12}{c|}}\hline
            \backslashbox{\bf Action}{\bf Score}
            &\textbf{\#1}&\textbf{\#2}&\textbf{\#3}&\textbf{\#4}&\textbf{\#5}&\textbf{\#6}&\textbf{\#7}&\textbf{\#8}&\textbf{\#9}&\textbf{\#10}&\textbf{\#11}&\textbf{\#12}\\\hline\hline
            \textbf{W-P} & 4 & 8 & 16 & 12 & - & -&-&-&-&-&-&-\\\hline
            \textbf{W-S} & 10 & 14 & 19 & 15 & 10 &-&-&-&-&-&-&-\\\hline
            \textbf{SS-P} &1 &1 & 6& 8& 4& 4& 4&3 &3 &1 &2 &4\\\hline
            \textbf{SS-S} &3 & 19& 19& 13 &20 &-&-&-&-&-&-&-\\\hline
             \end{tabular}
                      } 
     \caption{Details of abnormality score ranges in the {\ds} dataset}
     \label{tab:dataset-statistic}
\end{table}

\begin{figure}[h]
\begin{center}
    \includegraphics[width=1\linewidth]{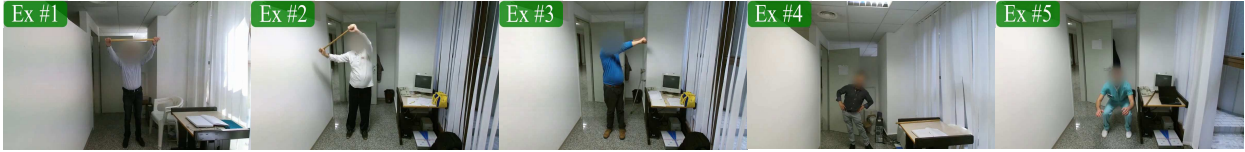}
    \end{center}
       \caption{{Sample frames of {\ki} for five different exercises.}}
    \label{fig:kimore sampels}
\end{figure}

\subsection{{\ki} \cite{capecci2019kimore}} \label{sec:kimore}
This is the only RGB single view rehabilitation movement dataset where the quality of movements have been annotated for quantitative scores. {\ki} \cite{capecci2019kimore} has 78 subjects (44 healthy, and 34 real patients suffering from Parkinson, Stroke, and back pain) performing five types of rehabilitation exercises (Ex \#1 to Ex \#5) for lower-back pain. All videos are frontal view - see sample frames in Fig. \ref{fig:kimore sampels}. 

{{\ki} \cite{capecci2019kimore} provides two types of scores, $PO_S$ and $CF_S$, with values in the range 0 to 50 for each exercise as defined by clinicians. $PO_S$ and $CF_S$ represent the motion of upper limbs and physical constraints during the exercise respectively.}

\section{Proposed Method}
\label{sec:proposed method}

\begin{figure*} [b]
\centering
\includegraphics[width=1.0\textwidth]{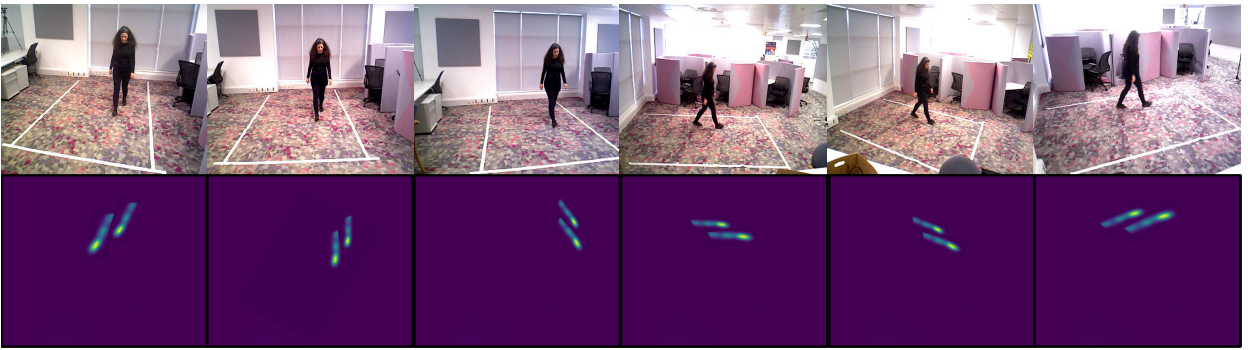} 
\vspace*{-1mm}
\caption{{Walking example - all six views, and corresponding {{trajectory maps}} for feet.} }
\label{fig:different apperance views}
\end{figure*}

Although its appearance changes significantly when we observe an instance of human movement from different viewpoints,  the 2D spatio-temporal trajectories generated by body joints in a sequence are affine transformations of each other. For example,
see Fig. \ref{fig:different apperance views}, where the {{trajectory maps of just the feet joints}} appear different in orientation, spatial location and scale. 
Thus, our hypothesis is that by extracting body joint {{trajectory maps}} that are 
translation, rotation, and scale invariant,  we {should be} able to assess the quality of movement from arbitrary viewpoints one may encounter in-the-wild.

{The proposed VI-Net network has a view-invariant trajectory descriptor module (VTDM) that feeds into a subsequent movement score module (MSM) as shown in Fig.~\ref{fig:overal schema}. In VTDM, first a 2D convolution filter is applied on stacked heatmaps of each body joint over the video clip frames to generate a trajectory descriptor {map}. Then, the Spatial Transformer Network (STN) \cite{jaderberg2015spatial} is applied to the trajectory descriptor to make it view-invariant. The spatio-temporal descriptors from all body joints are then stacked as input into the {MSM} module, which can be implemented by an adapted, pre-trained {CNN} to {learn} the relationship {amongst the joint trajectories} and provide a score for the overall quality of movement. We illustrate the flexibility of MSM by implementing two different pre-trained networks, VGG-19 and ResNeXt-50, and  compare their results. 
{VI-Net} is trained in an end-to-end manner. As the quality of movement scores in our {\ds} dataset are discrete, we use classification to obtain our predicted score. Table \ref{tab:network details} carries further details of our proposed VI-Net. }

\begin{table}[b]
\centering
\scalebox{1.0}{
    \begin{tabular}{|c|l||l|}           \hline
    \textbf{} & \multicolumn{1}{c||}{\textbf{VTDM}} & \multicolumn{1}{c|}{\textbf{MSM} (Adapted VGG-19 or ResNeXt-50)}\\ \hline \hline
    \multirow{5}{*}{\rotatebox{90}{\bf VI-Net}} &  \textbf{1st layer:} $\{C2(3 \times 3,T)\}\times 1$, BN, ReLU &  \textbf{1st layer VGG-19}: $\{C2(3 \times 3,J)\}\times 64$, BN, ReLU\\
    & \textbf{Localisation Network:}  & \textbf{1st layer ResNeXt-50:}  \\
	& $\{C2(5 \times 5,1)\}\times 10$,$\{MP(2\times 2)\}$, ReLU, & $\{C2(7 \times 7,J)\}\times 64$, $\{MP(3 \times 3)\}$, ReLU\\ 
    & $\{C2(5 \times 5,10)\}\times 10$,$\{MP(2 \times 2)\}$, ReLU, & \textbf{Middle layers:}  As in VGG-19/ResNeXt-50\\
    & $\{FC(32)\}$, ReLU, $\{FC(4)\}$ & \textbf{Last layer:} $\{FC(S+1)\}$\\\hline

        \end{tabular}
     }
\caption{{VI-Net's modules: $\{C2(d \times d,ch)\}\times n$: $n$ 2D convolution filters with size $d$ and $ch$ channel size, $MP(d \times d)$: 2D max pooling with size $d$, $FC(N)$: FC layer with $N$ outputs. $T$ is the  \# of clip frames, $J$ is the \# of joints and {$S$ is maximum score for a movement type}.}} 
\label{tab:network details}
\end{table}

{\bf Generating a Joint Trajectory Descriptor --} First, we extract human body joint heatmaps by estimating the probability of each body joint at each image pixel, per frame for a video clip {with}  $T$ frames, by applying OpenPose \cite{cao2017realtime}. 
{Even though it may seem that our claim to be an RGB-only method may be undermined by the use of a method which was built by using joint labelling, the fact remains that OpenPose is used in this work as an existing tool, with no further joint labelling or recourse to non-RGB data.}
{Other methods, \eg \cite{kocabas2019self}, which estimate body joint heatmaps from RGB images can equally be used}. 

To reduce computational complexity, 
we retain the first 15 joint heatmaps of the BODY-25 version of OpenPose. This is further
motivated by the fact, highlighted in \cite{paiement2014online}, that the remaining joints only provide repetitive information. Then, for each body joint $j\in \{1,2,...,J\!=\!15\}$, we stack its heatmaps over the $T$-frame video clip to get the 3D heatmap $J_j$ of size $W\!\times\!H\!\times\!T$ which then becomes the input {to} our VTDM module. 
To obtain {a body joint's} {trajectory descriptor} $\Lambda_j$, the processing in VTDM starts with the application of a convolution filter $\Phi$ on $J_j$ to aggregate its spatial poses over time, i.e.
\begin{equation}
    \Lambda_j=J_j*\Phi ~,
\end{equation}
where $\Lambda_j$ is of size $W\!\times\!H\!\times\!1$.
We experimented with both 2D and 3D convolutions, and  found that a $3\!\times\!3$ 2D convolution filter yields the best results. 

{\bf Forging a View-Invariant Trajectory Descriptor -- } 
In the next step of the VTDM module, {we experimented with STN \cite{jaderberg2015spatial}, DCN \cite{dai2017deformable,zhu2019deformable}, and ETN \cite{tai2019equivariant} networks, and found STN \cite{jaderberg2015spatial} the best performing option to forge a view-invariant trajectory descriptor out of $\Lambda_j$}.

STN  can be applied to feature maps of a CNN's layers to render {the output} translation, rotation, scale, and shear invariant. It is composed of three stages. {At first, a CNN-regression network, referred to as the localisation network, is applied to our joint trajectory descriptor $\Lambda_j$ to estimate the parameters for a 2D affine transformation matrix, $\theta=f_{loc}(\Lambda_j)$. Instead of the original CNN in \cite{jaderberg2015spatial}, which applied 32 convolution filters followed by two fully connected (FC) layers, we formulate our own localisation network made up of 10 convolution filters followed by two FC layers. The rationale for this is that our trajectory descriptor {maps} are not as complex as RGB images, and hence fewer filters are sufficient to extract their features. The details of our localisation network’s layers are provided in Table \ref{tab:network details}.} Then, in the second stage, to estimate each pixel value of our view-invariant trajectory descriptor $\bar\Lambda_j$, a sampling kernel is applied on specific regions of  $\Lambda_j$, where the centres of these regions are defined on a sampling grid. This sampling grid {${\Gamma_\theta}(G)$ is generated from a general grid $G=\{(x^{g}_i,y^{g}_i)\}, i\in\{1,\dots,W^{'}\!\times\!H^{'}\}$ and the predicted transformation parameters}, such that
\begin{equation}
 \label{eq:affine}
    \binom{x^{\Lambda_j}_i}{y^{\Lambda_j}_i}={{\Gamma_\theta}(G_i)}=\begin{bmatrix}
        \theta_{11} & \theta_{12} \\
        \theta_{21} & \theta_{22}
     \end{bmatrix}\times\binom{x^{g}_i}{y^{g}_i}~,
\end{equation}
where $\Gamma_\theta(G)=\{(x^{\Lambda_j}_i,y^{\Lambda_j}_i), i\in\{1,\dots,W^{'}\!\times\!H^{'}\}$ are {the centers of the regions of $\Lambda_j$ 
the sampling kernel is applied to, in order to  generate the new pixel values of the output feature map $\bar\Lambda_j$. Jaderberg \etal \cite{jaderberg2015spatial} recommend the use } of different types of transformations to generate the sampling grid  ${\Gamma_\theta}(G)$ based on the problem domain. In VTDM, we use the 2D affine transformations {shown} in Eq. \ref{eq:affine}.
Finally, the sampler takes both $\Lambda_j$ and ${\Gamma_\theta}(G)$ to generate a view-invariant trajectory descriptor $\bar\Lambda_j$ from $\Lambda_j$ at the grid points by bilinear interpolation. 

{\bf Assessing the Quality of Human Movement - }
In the final part of VI-Net (see Fig. \ref{fig:overal schema}), the collection of view-invariant trajectory descriptors $\bar\Lambda_j$ for joints  $j\in \{1,2,...,J\!=\!15\}$, are stacked  into a global descriptor $\bar\Lambda$ and passed through a  pre-trained network in the MSM module to assess the quality of movement of the joints. 
{VGG-19 and ResNeXt-50 were chosen for their state-of-the-art performances, popularity, and availability. 
For VGG-19, its first layer was replaced with a new 2D convolutional layer, including {$3\times3$ convolution filters} with channel size $J$ (instead of 3 used for RGB {input} images), and for  ResNeXt-50 its first layer was replaced with {$7\times7$ convolution filters} with channel size $J$.  The last FC layer in each case} was {modified} to allow movement quality scoring through classification where each score is considered  as a class, i.e. {for a movement with maximum score $S$}, where $S\!\!=\!\!4$ for W-P, $S\!\!=\!\!5$ for W-S and SS-S, and $S\!\!=\!\!12$ for SS-P movements. The last FC layer of VI-Net has {$S+1$}  output units. 

{Although VGG-19/ResNeXt-50 were trained on RGB images, we still benefit from their pretrained weights, since our new first layers were initialised with their original first layer weights. The output of this modified layer has the same size as the output of the layer it replaces (Table \ref{tab:network details}), so the new layer is compatible with the rest of network. In addition,  we normalize the pixel values of the trajectory heatmaps to be between 0 and 255, i.e. the same as RGB images on which VGG and ResNeXt were trained on, and trajectory descriptor maps have shape and  intensity variations - thus the features extracted from them would be as equally valid as for natural images on which VGG and ResNeXt operate.}


\section{Experiments and Results}
\label{sec:experimental results}
We {first} report on two sets of experiments {on {\ds}} to evaluate the performance of VI-Net to assess  quality of movement, based around cross-subject and cross-view scenarios. {Then, to show the efficiency of VI-Net on other datasets and movement types, we present its results also on the single view  {\ki} dataset.}
We used Pytorch on two GeForce GTX 750 GPUs. {All networks} 
were trained for 20 epochs using stochastic gradient descent optimization with initial learning rate of 0.001, and batch size 5. To evaluate the performance of the proposed method, we used Spearman's rank correlation {as used in} \cite{pan2019action, parmar2019and,li2018end}.

{\bf Dataset Imbalance --}
{It can be seen from Tables \ref{tab:dataset detalis} and \ref{tab:dataset-statistic} that the number of sequences for score $0$ (normal) is many more than the number of sequences for other individual scores, so we randomly selected 15 normal sequences for W-P, W-S, SS-S movements and 4 normal sequences for SS-P to mix with abnormal movements to perform all our experiments. To {further} address the imbalance,  we applied offline temporal cropping to add new sequences.}

\begin{figure} 
    \centering
    \includegraphics[width=0.9\textwidth]{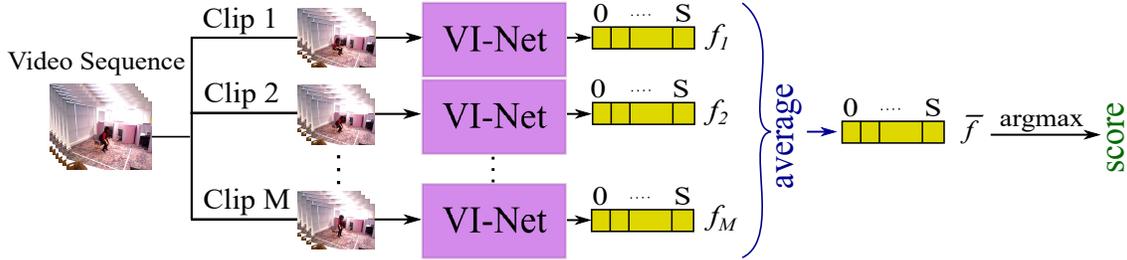}
    \caption{Scoring process for a full video sequence in testing phase.}
    \label{fig:testing phase}
\end{figure}

{{\bf {Network Training and Testing} -- }
In both the training and testing phases, video sequences were divided into 16-frame clips (without overlaps). In training, the clips were selected randomly from amongst all video sequences of the training set, and passed to VI-Net.  Then, the weights were updated following a {cross-entropy} loss, } 
\begin{equation}
\label{eq:C}
    {L_C(f,s)} = -{log(\frac{exp(f(s))}{\sum_{k=0}^{S}{exp(f(k))}})}~,
\end{equation}
where $f(.)$ is the $S+1$ dimensional output of the last fully connected layer and $s$ is the video clip's ground truth {label/score}. {In testing, every 16-frame clip of a video sequence was passed to VI-Net. After averaging the outputs of the last fully connected layer across each class for all the clips, we then set the score for the whole video sequence as the maximum of the clip scores (see Fig. \ref{fig:testing phase}), i.e. } 
\begin{equation}
\label{}
    {s} = \operatorname*{argmax}_k ({\bar{f}(k)}=\frac{1}{M}\sum_{m=1}^{M}{f_m(k)})~,
\end{equation}
where $k\in\{0,1,\dotsc {S}\}$ and $M$ is the number of clips for a video. 

{\bf Comparative Evaluation --}
As we are not aware of any other RGB-based view-invariant method to assess quality of movement, we are unable to compare VI-Net's performance to other methods under a cross-view scenario. However, for cross-subject {and single-view scenarios}, we evaluate against (a) a C3D baseline (fashioned after Parmar and Morris \cite{parmar2019and}) by combining the outputs of the C3D network to score a sequence in the test phase in the same fashion as in VI-Net, and (b) the pre-trained, {fine-tuned} I3D \cite{carreira2017quo}.  We also provide an ablation study for all scenarios by removing STN from VI-Net to analyse the impact of this part of the proposed method.

\subsection{{Cross-Subject} Quality of  Movement Analysis}
In this experiment, all available views were used in both training and testing, while the subjects performing the actions were {distinct. We applied} $k$-fold cross validation where $k$ is the number of scores for each movement. 
Table \ref{tab:multi-view results} shows that VI-Net outperforms networks based on C3D (after  \cite{parmar2019and}) and I3D \cite{carreira2017quo} for all types of movements, regardless of whether VGG-19 or ResNeXt-50 are used in the MSM module. 
{While I3D results are mostly competitive, C3D performs less well due to its shallower nature, and larger number of parameters, exacerbated by \ds's relatively small size. {We show in Section \ref{sec:single-view analysis} that C3D performs significantly better on a larger dataset}}. 

As ablation analysis, to test the effectiveness of STN, we present VI-Net's results with and without STN in Table \ref{tab:multi-view results}. It can be observed that the improvements with STN are not necessarily consistent across the actions {since when all viewpoints are used in training,} the MSM module gets trained on all {trajectory orientations} {such that the effect of STN is often overridden}. 

 

\begin{table}[t!]
    \centering
        \scalebox{1.0}{
            \begin{tabular}{|l|c|c|c|c|c|c|c|}\hline
            \multicolumn{3}{|c|}{\backslashbox{\bf Method}{\bf Action}}
            &\textbf{W-P}&\textbf{W-S}&\textbf{SS-P}&\textbf{SS-S}&\textbf{Avg}\\\hline\hline
           \multicolumn{3}{|c|}{\textbf{Custom-trained C3D (after \cite{parmar2019and})}}  & {0.50} & {0.37} & {0.25} & {0.54} & \cellcolor{black!10}{0.41}\\\hline
           \multicolumn{3}{|c|}{\textbf{Pre-trained I3D} {\bf \cite{carreira2017quo}}} & {0.79} &{0.47}& {0.54} &{0.55} & \cellcolor{black!10}{0.58}\\ \hline
           \multirow{4}{*}{\textbf{VI-Net}}&\multirow{2}{*}{\textbf{VTDM+MSM (VGG-19)}}&\textbf{{ w/o STN}}& 0.81 & 0.49 & 0.57 & {\bf 0.74} & \cellcolor{black!10}{0.65}\\\cline{3-8}
            &&\textbf{w STN} &0.82 &{ 0.52} & 0.55 &{0.73 }& \cellcolor{black!10}{0.65} \\\cline{2-8}
           &\multirow{2}{*}{\textbf{VTDM+MSM (ResNeXt)}}&{ \textbf {w/o STN}} & {\bf 0.87} & {\bf 0.56} & 0.48 &{0.72}  & \cellcolor{black!10}{0.65}\\\cline{3-8}
            &&\textbf{w STN} &{\bf 0.87} &{ 0.52} & {\bf 0.58} &{ 0.69}& \cellcolor{black!10}{\bf 0.66} \\\hline
            
             \end{tabular}
                      } 
     \caption{{Comparative cross-subject results on {\ds}.}}
     \label{tab:multi-view results}
\end{table}


\subsection{{Cross-View} Quality of Movement Analysis}
{We evaluate the generalization ability of VI-Net on unseen views by using cross-view scenarios, 
i.e. distinct training and testing views of the scene, while data from all subjects is utilised.
{{Recent works such as \cite{wang2018dividing,li2018unsupervised,lakhal2019view,li2017domain}, provide cross-view results only when their network is trained on multiple views. As recently noted by Varol et al. \cite{varol19_surreact}, a highly challenging scenario} in view-invariant action recognition would be to obtain cross-view results by training from only one viewpoint. Therefore, } we performed the training and testing for each movement type such that (i) we trained from one view only and tested on all other views (as reasoned in Section \ref{sec:intro}), and in the next experiment, (ii) we trained on a combination of one frontal view (views 1 to 3) and one side view (views 4 to 6) and tested on all other available views. Since for the latter case there are many combinations, we  show results for only selected views: 
view 2 $\approx0^\circ$ with all side views, and view 5 $\approx90^\circ$ with all frontal views.}

{Since in cross-view analysis all subjects are used in both training and testing, applying the C3D and I3D models would be redundant because they would simply learn the appearance and shape features of our participants in our study and their results would be unreliable.} 

{In {\ds}, when observing a movement from the frontal views, there is little or almost no occlusion of {relevant} body parts. However, when observing from side views, occlusions {resulting from missing or noisy joint heatmaps from OpenPose,} can occur for a few seconds or less (short-term), or for almost the whole sequence (long-term).}
{Short term occlusions are more likely in walking movements W-P and W-S, while long-term occlusions occur more often in sit-to-stand movements ({SS-P} and {SS-S}).} 

{The results of our view-invariancy experiments, using single views only in training, are shown in Table \ref{tab:unseen single-views-result-1}.  It can be seen  that for walking movements W-P and {W-S}, VI-Net is able to assess the movements from unseen views well, with the best results reaching $0.73$ and $0.66$ rank correlation respectively (yellow highlights), and only relatively affected by short term occlusions. However, for sit-to-stand movements {SS-P} and SS-S, the long-term occlusions during these movements affect the integrity of the trajectory descriptors and the performance of VI-Net is not as strong, with the best results reaching $0.52$ and $0.56$ respectively (orange highlights).} {Note, for all action types, when VI-Net has STN with adapted ResNeXt, it performs best on average.} 


Table \ref{tab:unseen two-views-result-1} shows the results for each movement type when one side view and one frontal view are combined for training. VI-Net's performance improves compared to the single view experiment above {with the best results reaching $0.92$ and $0.83$ for W-P and W-S movements (green highlights) and $0.61$ and $0.64$ for SS-P and SS-S movements (purple highlights)}, because the network is effectively trained with both short-term and long-term occluded trajectory descriptors.
These results also show that 
on average VI-Net performs better with adapted ResNeXt-50  for walking movements ({W-P} and {W-S}) and with adapted VGG-19 for sit-to-stand movements ({SS-P} and {SS-S}). {This is potentially because  ResNext-50's variety of filter sizes are better suited to the variation in 3D spatial changes of joint trajectories inherent in walking movements compared to VGG-19's $3\times 3$ filters which can tune better to the more spatially restricted sit-to-stand movements}. We also note that {the fundamental purpose of STN in VI-Net is to ensure efficient cross-view performance is possible when the network is trained from a single view only. It would therefore be expected and plausible that STN's effect would diminish as more views are used since the MSM module gets trained on more trajectory orientations (which we verified experimentally by training with multiple views)}. 

\begin{table}[h]
\centering
\scalebox{0.9}{
     \begin{tabular}{|c|c|c|c|c|c||c|c|c|c|c|c|} \hline 
    & \multirow{3}{*}{\textbf{View}}&\multicolumn{2}{c|}{\textbf{VTDM+MSM}}&\multicolumn{2}{c||}{\textbf{VTDM+MSM}} & \textbf{}&\multirow{3}{*}{\textbf{View}}&\multicolumn{2}{c|}{\textbf{VTDM+MSM}}&\multicolumn{2}{c|}{\textbf{VTDM+MSM}}\\
    \textbf{}& & \multicolumn{2}{c|}{\textbf{(VGG-19)}}& \multicolumn{2}{c||}{\textbf{(ResNeXt-50)}}& \textbf{}& &\multicolumn{2}{c|}{\textbf{(VGG-19)}}&\multicolumn{2}{c|}{\textbf{(ResNeXt-50)}}\\\cline{3-6} \cline{9-12}
    \textbf{}&&\textbf{w/o STN}&\textbf{w STN}&\textbf{w/o STN}&\textbf{w STN} & \textbf{}&&\textbf{w/o STN}&\textbf{w STN}&\textbf{w/o STN}&\textbf{w STN}\\ \hline \hline
	\multirow{7}{*}{\rotatebox{90}{W-P}}&{1}&{0.51} & {\bf 0.67}&  {0.64} &  {\bf 0.67}  &  \multirow{7}{*}{\rotatebox{90}{W-S}}&{1}&0.51 &0.43 &0.60  & {\bf 0.64}    \\ \cline{2-6} \cline{8-12}	         
	&{2}&0.69 &0.66 & 0.58 &  {\bf0.72}                  & 	                                    &{2} &0.47 &0.54  &0.55  & {\bf 0.62}   \\ \cline{2-6} \cline{8-12}                               
	&{3}&0.62 &0.66 & 0.63 &  {\bf0.70}                  &                                        &{3} & {\bf0.64}&0.56 & {0.61}&  0.59       \\ \cline{2-6} \cline{8-12}          	                 
	&{4}&0.67 &0.64  &  {\bf 0.72} &   {\bf 0.72}        &                                        &{4}& 0.60&0.59  &  0.60&  \cellcolor{yellow}  {\bf 0.66} \\ \cline{2-6} \cline{8-12}                                        
	&{5}& 0.67& 0.67& 0.68 &   {\bf 0.71}                &                                       	&{5}&0.62 & 0.60 &  0.62&    {\bf 0.63} \\ \cline{2-6} \cline{8-12}                                        
	&{6}& 0.69&0.72 & 0.69& \cellcolor{yellow} {\bf 0.73}                &                                        &{6} & 0.46& 0.40&  0.53&   {\bf 0.60}  \\ \cline{2-6} \cline{8-12}                                         
	&\cellcolor{black!10}{\bf Avg }&\cellcolor{black!10} 0.64 & \cellcolor{black!10} 0.67&  \cellcolor{black!10} 0.65 &  \cellcolor{black!10} {\bf 0.70} & &\cellcolor{black!10}{\bf Avg }&\cellcolor{black!10} 0.55 & \cellcolor{black!10}0.52 &  \cellcolor{black!10} 0.58 & \cellcolor{black!10} {\bf 0.62}     \\ \hline \hline

	\multirow{7}{*}{\rotatebox{90}{SS-P}}&{1}& 0.30 & {\bf 0.32} &  { 0.25}&  {0.25}  &                  \multirow{7}{*}{\rotatebox{90}{SS-S}}&{1}&0.36 & {\bf 0.49} &  0.44& { 0.45}   \\ \cline{2-6} \cline{8-12}                   	
	&{2}& 0.27 & 0.31 & 0.31 &  {\bf 0.32}                       &                  &{2}&0.47 & 0.40&  \cellcolor{orange!70} {\bf 0.56} &  \cellcolor{orange!70} {\bf 0.56}                 \\ \cline{2-6} \cline{8-12}                        
	&{3}&0.16 & 0.23 &  0.36&    {\bf 0.43}               &                  &{3}& 0.37& {\bf 0.52}&  0.38&   { 0.43}                      \\ \cline{2-6} \cline{8-12}                      
	&{4}& 0.10 & 0.34  &  0.44&   {\bf 0.49}               &                  &{4}&0.38 & 0.34 &  0.41 &   {\bf 0.54}                      \\ \cline{2-6} \cline{8-12}                       
	&{5}&0.50 & \cellcolor{orange!70} {\bf 0.52}  &  { 0.43} &    0.45             &                  &{5}& 0.26&  {\bf 0.50} & {\bf 0.50} &    0.48                       \\ \cline{2-6} \cline{8-12}                      
	&{6}&0.41 &  0.24 &  {\bf 0.48}&    0.44              &                  &{6}&0.21 & {\bf 0.28} &  0.13&    { 0.16}                        \\ \cline{2-6} \cline{8-12}                  
	&\cellcolor{black!10}{\bf Avg }&\cellcolor{black!10} 0.29 & \cellcolor{black!10} 0.32 & \cellcolor{black!10} {0.37} &  \cellcolor{black!10} {\bf 0.39} & &\cellcolor{black!10}{\bf Avg} &\cellcolor{black!10} 0.34 & \cellcolor{black!10} 0.42& \cellcolor{black!10} 0.40 & \cellcolor{black!10} {\bf 0.43}        \\  \hline 

        \end{tabular}
        }
    \caption{{Cross-view results for all {actions} with single-view training.}} 
    \label{tab:unseen single-views-result-1} 
\end{table}

\begin{table}[h]
\centering
\scalebox{0.91}{
    \begin{tabular}{|c|c|c|c|c|c||c|c|c|c|c|c|} \hline 
    & \multirow{3}{*}{\textbf{View}}&\multicolumn{2}{c|}{\textbf{VTDM+MSM}}&\multicolumn{2}{c||}{\textbf{VTDM+MSM}} & \textbf{}&\multirow{3}{*}{\textbf{View}}&\multicolumn{2}{c|}{\textbf{VTDM+MSM}}&\multicolumn{2}{c|}{\textbf{VTDM+MSM}}\\
    \textbf{}& & \multicolumn{2}{c|}{\textbf{(VGG-19)}}& \multicolumn{2}{c||}{\textbf{(ResNeXt-50)}}& \textbf{}& &\multicolumn{2}{c|}{\textbf{(VGG-19)}}&\multicolumn{2}{c|}{\textbf{(ResNeXt-50)}}\\\cline{3-6} \cline{9-12}
    \textbf{}&&\textbf{w/o STN}&\textbf{w STN}&\textbf{w/o STN}&\textbf{w STN} & \textbf{}&&\textbf{w/o STN}&\textbf{w STN}&\textbf{w/o STN}&\textbf{w STN}\\ \hline \hline
	\multirow{6}{*}{\rotatebox{90}{W-P}}&{2,4}& 0.77 & 0.81 & 0.87&  {\bf 0.89}  &  \multirow{6}{*}{\rotatebox{90}{W-S}}&{2,4}& 0.58& 0.72 & {\bf 0.81} &  0.73    \\ \cline{2-6} \cline{8-12}	         
	&{2,5}& 0.72 & 0.75 & 0.90 &    \cellcolor{green!40}{\bf 0.92}                   &       &{2,5}& {0.74} & {0.74} & {0.80} & {\bf 0.81} \\ \cline{2-6} \cline{8-12}                    
	&{2,6}& 0.75& 0.76& 0.73 &     {\bf 0.77}                  &      &{2,6}& {0.64} & {0.67} & {\bf 0.74} &  {0.68}  \\ \cline{2-6} \cline{8-12}                             
	&{1,5}& {0.70} & {0.76} & {\bf 0.80}  &   0.75                  &     &{1,5}& {0.70} & {0.68} & \cellcolor{green!40}{\bf 0.83} &  {0.81}    \\ \cline{2-6} \cline{8-12}                               
	&{3,5}& 0.73& 0.79&  {\bf 0.87} &      0.84                &      &{ 3,5}& {0.66} & {0.66} & {\bf 0.82} &  {0.79 }  \\ \cline{2-6} \cline{8-12}                            
	&\cellcolor{black!10}{\bf Avg }&\cellcolor{black!10} 0.73&\cellcolor{black!10} 0.77& \cellcolor{black!10} {\bf 0.83} & \cellcolor{black!10}  {\bf 0.83}& &\cellcolor{black!10}{\bf Avg }&\cellcolor{black!10} 0.66& \cellcolor{black!10}0.69 &\cellcolor{black!10}{\bf 0.80} &\cellcolor{black!10} {0.76}  \\ \hline \hline 
                                       
	\multirow{6}{*}{\rotatebox{90}{SS-P}}&{2,4}& {\bf 0.55}&0.52 & {0.41} & { 0.46}  &  \multirow{6}{*}{\rotatebox{90}{SS-S}}& {2,4}&0.57 &{\bf 0.64}  & 0.54 &{\bf 0.64}\\ \cline{2-6} \cline{8-12}                   	
	&{2,5}& {\bf 0.60} & 0.53& 0.49 & 0.46                        &    &{2,5}&0.62 &0.56 & {\bf 0.63}& 0.61\\ \cline{2-6} \cline{8-12}      
	&{2,6}&{\bf 0.48} & 0.35& 0.36 & 0.42                     &    &{2,6}& 0.50& {\bf 0.62} & {0.48} &  0.46\\ \cline{2-6} \cline{8-12}
	&{1,5}&0.46 & {\bf 0.55}& 039 & 0.52                  &    & {1,5}&{\bf 0.64} &0.53 & 0.48 &  0.58       \\ \cline{2-6} \cline{8-12}        
	&{3,5}& \cellcolor{blue!30}{\bf 0.61}& 0.40& 0.43 & 0.47                       &    &{3,5}& 0.62& 0.60 & 0.63& \cellcolor{blue!30}{\bf 0.67}       \\ \cline{2-6} \cline{8-12}        
	&\cellcolor{black!10}{\bf Avg }& \cellcolor{black!10} {\bf 0.54}& \cellcolor{black!10}0.47 & \cellcolor{black!10} {0.41} &    \cellcolor{black!10} { 0.46}& &\cellcolor{black!10}{\bf Avg }& \cellcolor{black!10} {\bf 0.59}& \cellcolor{black!10} {\bf 0.59} &\cellcolor{black!10} 0.55 & \cellcolor{black!10}  { 0.58} \\ \hline 

        \end{tabular}
        }
    \caption{{Cross-view results for all {actions} with two-view training.}} 
    \label{tab:unseen two-views-result-1} 
\end{table}


\subsection{{Single-View} Quality of Movement Analysis}
\label{sec:single-view analysis}
{Next, we provide the results of VI-Net on the single-view {\ki} dataset, to illustrate that it can be applied to such data too. {\ki} provides two types of scores, $PO_S$ and $CF_S$ (see Subsection \ref{sec:kimore}) which have a strong correspondence to each other, such that if one is low for a subject, so is the other. Hence, we trained the network based on a single, summed measure to predict a final score ranging between 0 and 100 for each action type. We include $70\%$ of the subjects for training and retain the remaining $30\%$ for testing ensuring each set contains a balanced variety of scores from low to high.} 

{Table \ref{tab:single-view results} shows the results of C3D baseline (after \cite{parmar2019and}), pre-trained, fine-tuned I3D \cite{carreira2017quo} and VI-Net on {\ki}. 
It can be seen that VI-Net outperforms the other methods for all movement types except for Exercise \#3. {VI-Net with adapted VGG-19 performs better than with ResNeXt-50 for all movement types.} {This may be because, similar to sit-to-stand movements in {\ds}, where VI-Net performs better with VGG-19, all movements types in {\ki} are also performed at the same location and distance from camera, and thus carry less variation in 3D trajectory space. This shows that our results are consistent in this sense across both datasets.} }

{In addition, although all sequences in both training and testing sets have been captured from the same view, VI-Net's performance on average improves with STN. This can be attributed to STN improving the network generalization on different subjects. Also, unlike in {\ds}'s cross-subject results where C3D performed poorly, the results on {\ki} for C3D are promising because {\ki} has more data to help the network train more efficiently.}

\begin{table}[h]
    \centering
        \scalebox{1.0}{
        \begin{tabular}{|c|c|c|c|c|c|c|c|c|}\hline
            \multicolumn{3}{|c|}{\backslashbox{\bf{Method}}{\bf{Action}}} &\textbf{Ex {\#}1}&\textbf{Ex {\#}2}&\textbf{Ex {\#}3}&\textbf{Ex {\#}4}& \textbf{Ex {\#}5} &\textbf{Average}\\\hline\hline
             \multicolumn{3}{|c|}{\textbf{Custom-trained C3D (after \cite{parmar2019and})}} & { 0.66} & { 0.64} & {\bf 0.63} & {0.59} & {0.60} & \cellcolor{black!10}{ 0.62}\\\hline
            \multicolumn{3}{|c|}{\textbf{Pre-trained I3D} {\bf \cite{carreira2017quo}}} & {0.45} & {0.56} & {0.57} & { 0.64} & {0.58} & \cellcolor{black!10}{0.56} \\\hline
          \multirow{4}{*}{\textbf{VI-Net}}&\multirow{2}{*}{\textbf{VTDM+MSM (VGG-19)}}&\textbf{ w/o STN} & {0.63} & {0.50} & {0.55} & {\bf 0.80} & { \bf 0.76} & \cellcolor{black!10}{\bf} 0.64 \\\cline{3-8}
          &&\textbf{ w STN} & {\bf 0.79} & {\bf 0.69} & {0.57} & {0.59} & {0.70} & \cellcolor{black!10}{\textbf {0.66}} \\ \cline{2-8}
          &\multirow{2}{*}{\textbf{VTDM+MSM (ResNeXt-50)}}&\textbf{ w/o STN} & {0.55} & {0.42} & {0.33} & {0.62} & {0.57} & \cellcolor{black!10}{0.49} \\\cline{3-8}
          & &\textbf{w STN} & {0.55} & {0.62} & {0.36} & {0.58} & { 0.67} & \cellcolor{black!10}{0.55} \\\hline
             \end{tabular}
            }
            \caption{{Comparative results} on the single-view {\ki} dataset.}
     \label{tab:single-view results}
\end{table}

\section{Conclusion}
\label{sec:concolusion}
{View-invariant human movement analysis from RGB is a significant challenge in {action analysis applications, such as sports, skill assessment, and healthcare monitoring}.}
In this paper, we proposed a novel {RGB based} view-invariant method to assess the quality of human movement {which can be trained from a relatively small dataset} and without any knowledge about viewpoints used for data capture. We also introduced  {\ds}, the only multi-view, non-skeleton, non-mocap, rehabilitation movement dataset to evaluate the performance of the proposed method, which may also serve well for comparative analysis for the community. We demonstrated that the proposed method is applicable to  cross-subject, cross-view, and single-view movement analysis by achieving average rank correlation 0.66 on cross-subject and 0.65 on unseen views when trained from only two views, and 0.66 on single-view setting. 

{\bf {~\\Acknowledgements} -- } 
The 1st author is grateful to the University of Bristol for her scholarship. The authors would also like to thank Dr Alan Whone and Dr Harry Rolinski of Southmead Hospital - clinical experts in Neurology - for fruitful discussions on the QMAR dataset. This work was performed under the SPHERE Next Steps Project funded by the UK Engineering and Physical Sciences Research Council (EPSRC), Grant EP/R005273/1.

\bibliographystyle{unsrt} 
\bibliography{vinet}
\end{document}